# Solving Highly Constrained Search Problems
# with Quantum Computers

**Tad Hogg**                                                HOGG@PARC.XEROX.COM
*Xerox Palo Alto Research Center*
*3333 Coyote Hill Road Palo Alto, CA 94304 USA*

## Abstract

A previously developed quantum search algorithm for solving 1-SAT problems in a single step is generalized to apply to a range of highly constrained $k$-SAT problems. We identify a bound on the number of clauses in satisfiability problems for which the generalized algorithm can find a solution in a constant number of steps as the number of variables increases. This performance contrasts with the linear growth in the number of steps required by the best classical algorithms, and the exponential number required by classical and quantum methods that ignore the problem structure. In some cases, the algorithm can also guarantee that insoluble problems in fact have no solutions, unlike previously proposed quantum search algorithms.

## 1. Introduction

Quantum computers (Benioff, 1982; Bernstein & Vazirani, 1993; Deutsch, 1985, 1989; Di-Vincenzo, 1995; Feynman, 1986; Lloyd, 1993) offer a new approach to combinatorial search problems (Garey & Johnson, 1979) with *quantum parallelism*, i.e., the ability to operate simultaneously on many classical search states, and *interference* among different paths through the search space. A quantum algorithm to rapidly factor integers (Shor, 1994), a problem thought to be intractable for classical machines, offers a dramatic example of how these features of quantum mechanics can be exploited.

While several additional algorithms have been developed (Boyer, Brassard, Hoyer, & Tapp, 1996; Cerny, 1993; Grover, 1997b, 1997a; Hogg, 1996, 1998a; Terhal & Smolin, 1997), the extent to which quantum searches can improve on heuristically guided classical search methods remains an open question. Quantum algorithms can be based directly on classical heuristics, achieving a search cost that is the square root of the corresponding classical method (Brassard, Hoyer, & Tapp, 1998; Cerf, Grover, & Williams, 1998). Obtaining further improvement requires uniquely quantum mechanical methods. Heuristics exploit the structure of the search problems to greatly reduce the search cost in many cases. The success of these heuristics raises the question of whether the structure of search problems can form the basis of even better quantum algorithms. A suggestion that this is possible has been observed empirically for highly constrained problems (Hogg, 1998a), but the complexity of the algorithm precluded a definitive theoretical analysis of its behavior.

This paper presents a new quantum search algorithm that is extremely effective for some highly constrained search problems. These constraints also allow for effective classical heuristics, i.e., these problems are relatively easy. However, the new quantum algorithm requires even fewer steps than the best classical methods, providing another example of search problems for which quantum computers can outperform classical ones. More significantly, this algorithm illustrates how knowledge of the structure inherent in search problems can be used to develop new algorithms. Finally, because of its simplicity, the algorithm's behavior can be readily characterized analytically in some cases, conclusively demonstrating its asymptotic performance behavior in those cases.

Specifically the following two sections briefly review the ingredients of quantum programs and the satisfiability problem. The quantum algorithm for a particularly simple case is described in Section 4 and generalized in Section 5. The new algorithm is then evaluated for a variety of highly constrained problems. Finally some open issues are discussed, including a variety of ways this approach can be extended.

## 2. Quantum Computers

The state of a classical computer can be described by a string of bits, each of which is implemented by some two-state device. Quantum computers use physical devices whose full quantum state can be controlled. For example (DiVincenzo, 1995), an atom in its ground state could represent a bit set to 0, and an excited state for 1. The atom can be switched between these states and also be placed in a uniquely quantum mechanical *superposition* of these values, which can be denoted as a vector $\begin{pmatrix} \psi_0 \\ \psi_1 \end{pmatrix}$, with a component (called an *amplitude*) for each of the corresponding classical states for the system. These amplitudes are complex numbers. A superposition should not be confused with a probabilistic representation of ignorance about whether a classical bit is really 0 or 1. Nor is a superposition simply in between a 0 or 1, as could be the case with a 3 volt value for classical bits implemented as 0 and 5 volts. Instead, a superposition has no complete classical analog.

In contrast to a classical machine which, at any given step of its program, has a definite value for each bit, a quantum machine with $n$ quantum bits exists in a general superposition of the $2^n$ classical states for $n$ bits, described by the vector

$$\psi = \begin{pmatrix} \psi_0 \\ \vdots \\ \psi_{2^n-1} \end{pmatrix} \tag{1}$$

The amplitudes have a physical interpretation: when the computer's state is measured, the superposition randomly changes to one of the classical states with $|\psi_s|^2$ being the probability to obtain the state $s$. Thus amplitudes satisfy the normalization condition $\sum_s |\psi_s|^2 = 1$. This measurement operation is used to obtain definite results from a quantum computation.

Using this rich set of states requires operations that can rapidly manipulate the amplitudes in a superposition. Because quantum mechanics is linear and the normalization condition must always be satisfied, these operations are limited to unitary linear operators (Hogg, 1996). That is, a state vector $\psi$ can only change to a new vector $\psi'$ related to the original one by a unitary transformation, i.e., $\psi' = U\psi$ where $U$ is a unitary matrix[1] of dimension $2^n \times 2^n$. In particular, this requires that the operations be reversible: each output is the result of a single input. In spite of the exponential size of the matrix, in many cases the operation can be performed in a time that grows only as a polynomial in $n$ by quantum computers (Boyer et al., 1996; Hoyer, 1997). Importantly, the quantum computer does not explicitly form, or store, the matrix $U$. Rather it performs a series of elementary operations whose net effect is to produce the new state vector $\psi'$. On the other hand, the components of the new vector are not directly accessible: rather they determine the probabilities of obtaining various results when the state is measured.

---

1. A complex matrix $U$ is unitary when $U^\dagger U = I$, where $U^\dagger$ is the transpose of $U$ with all elements changed to their complex conjugates. Examples include permutations, rotations and multiplication by phases (complex numbers whose magnitude is one).

Important examples of such operations are reversible classical programs (Bennett & Landauer, 1985; Feynman, 1996). Let $P$ be such a program. Then for each classical state $s$, i.e., a string of bit values, it produces an output $s' = P[s]$, and each output is produced by only a single input. A simple example is a program operating with two bits that replaces the first value with the exclusive-or of both bits and leaves the second value unchanged, i.e., $P[00] = 00$, $P[01] = 11$, $P[10] = 10$ and $P[11] = 01$. When used with a quantum superposition, such classical programs operate independently and simultaneously on each component to give a new superposition. That is, a program operating with $n$ bits gives

$$P\left[\begin{pmatrix} \psi_0 \\ \vdots \\ \psi_{2^n-1} \end{pmatrix}\right] \rightarrow \begin{pmatrix} \psi'_0 \\ \vdots \\ \psi'_{2^n-1} \end{pmatrix} \tag{2}$$

where $\psi'_{s'} = \psi_s$ with $s' = P[s]$. This *quantum parallelism* allows a machine with $n$ bits to operate simultaneously with $2^n$ different classical states.

Unitary operations can also mix the amplitudes in a state vector. An example for $n = 1$ is

$$\frac{1}{\sqrt{2}} \begin{pmatrix} 1 & 1 \\ 1 & -1 \end{pmatrix} \tag{3}$$

This converts $\begin{pmatrix} 1 \\ 0 \end{pmatrix}$, which could correspond to an atom prepared in its ground state, to $\frac{1}{\sqrt{2}} \begin{pmatrix} 1 \\ 1 \end{pmatrix}$, i.e., an equal superposition of the two states. Since amplitudes are complex numbers, such mixing can combine amplitudes to leave no amplitude in some of the states. This capability for interference (Bernstein & Vazirani, 1993; Feynman, 1985) distinguishes quantum computers from probabilistic classical machines.

## 3. The Satisfiability Problem

NP search problems have exponentially many possible states and a procedure that quickly checks whether a given state is a solution (Garey & Johnson, 1979). Constraint satisfaction problems (CSPs) (Mackworth, 1992) are an example. A CSP consists of $n$ variables, $V_1, \ldots, V_n$, and the requirement to assign a value to each variable to satisfy given constraints. An *assignment* specifies a value for each variable.

One important CSP is the satisfiability problem (SAT), which consists of a logical propositional formula in $n$ variables and the requirement to find a value (true or false) for each variable that makes the formula true. This problem has $N = 2^n$ assignments. For $k$-SAT, the formula consists of a conjunction of clauses and each clause is a disjunction of $k$ variables, any of which may be negated. For $k \geq 3$ these problems are NP-complete. An example of such a clause for $k = 3$, with the third variable negated, is $V_1$ OR $V_2$ OR (NOT $V_3$), which is false for exactly one assignment for these variables: $\{V_1 = \text{false}, V_2 = \text{false}, V_3 = \text{true}\}$. A clause with $k$ variables is false for exactly one assignment to those variables, and true for the other $2^k - 1$ choices. Since the formula is a conjunction of clauses, a solution must satisfy every clause. We say an assignment conflicts with a particular clause when the values the assignment gives to the variables in the clause make the clause false. For example, in a four variable problem, the assignment

$$\{V_1 = \text{false}, V_2 = \text{false}, V_3 = \text{true}, V_4 = \text{true}\}$$

conflicts with the $k = 3$ clause given above, while

$$\{V_1 = \text{false}, V_2 = \text{false}, V_3 = \text{false}, V_4 = \text{true}\}$$

does not. Thus each clause is a constraint that adds one conflict to all assignments that conflict with it. The number of distinct clauses $m$ is then the number of constraints in the problem.

The assignments for SAT can also be viewed as bit strings with the correspondence that the $i^{th}$ bit is 0 or 1 according to whether $V_i$ is assigned the value false or true, respectively. In turn, these bit strings are the binary representation of integers, ranging from 0 to $2^n - 1$. For definiteness, we arbitrarily order the bits so that the values of $V_1$ and $V_n$ correspond, respectively, to the least and most significant bits of the integer. For example, the assignment

$$\{V_1 = \text{false}, V_2 = \text{false}, V_3 = \text{true}, V_4 = \text{false}\}$$

corresponds to the integer whose binary representation is 0100, i.e., the number 4.

For bit strings $r$ and $s$, let $|s|$ be the number of 1-bits in $s$ and $r \wedge s$ the bitwise AND operation on $r$ and $s$. Thus $|r \wedge s|$ counts the number of 1-bits both assignments have in common. We also use $d(r, s)$ as the Hamming distance between $r$ and $s$, i.e., the number of positions at which they have different values. These quantities are related by

$$d(r, s) = |r| + |s| - 2|r \wedge s| \tag{4}$$

An example 1-SAT problem with $n = 2$ is the propositional formula (NOT $V_1$) AND (NOT $V_2$). This problem has a unique solution: $\{V_1 = \text{false}, V_2 = \text{false}\}$, an assignment with the bit representation 00. The remaining assignments for this problem have bit representations 01, 10, and 11.

## 3.1 Highly Constrained SAT Problems

In general, SAT problems are difficult to solve. However, in a few simple cases the very regular structure of the search space allows much more effective algorithms. One example is given by 1-SAT problems. In this case, each clause eliminates one value for a single variable allowing classical algorithms to examine the variables independently, giving an overall search cost of $O(n)$ in spite of the exponentially large number of assignments. A 1-SAT problem has a solution if and only if each of the $m$ clauses involves a distinct variable. Otherwise, both values for some variable will be in conflict, i.e., making a clause false, resulting in no solutions.

This simple structure allows for rapid search. SAT problems with larger clauses have a more complicated structure. Nevertheless, when the $k$-SAT problems are highly constrained, their structure is close to that of 1-SAT. To see this, consider a soluble $k$-SAT problem. With respect to a particular solution of that problem, define the *good* value for each variable as the value (true or false) it is assigned in that solution, while the opposite value is the *bad* one for that variable. For $k$-SAT problems with many constraints, the *number* of bad values in an assignment can usually be determined rapidly from its number of conflicts, even though determining exactly *which* variables have incorrect assigned values requires first finding the solution. In such cases, using the number of bad values results in a tractable algorithm as long as *a priori* knowledge of the solution is not assumed.

For example, consider soluble problems with the largest possible number of constraints. For $k$-SAT, these maximally constrained soluble problems have $m = m_{\max}$ where

$$m_{\max} = \binom{n}{k}(2^k - 1) \tag{5}$$

i.e., the single solution precludes any clause that conflicts with it.

An assignment with $j$ bad values contains $\binom{n-j}{k}$ sets of $k$ variables all of which have the same values as the solution. Each of the remaining sets conflicts with at least one clause in the problem. Thus each assignment with $j$ bad values has

$$c_{\max}(j) = \binom{n}{k} - \binom{n-j}{k} \tag{6}$$

conflicts. This quantity grows strictly monotonically with $j$ for $j \leq n - k$, so in these cases $j$ is directly determined by the number of conflicts. Assignments with $n - k + 1 \leq j \leq n$ are not distinguishable.

Assignments with $j = n - k + 1$ can also be correctly determined by including neighborhood information. To see this, consider an assignment $s$ with $j$ bad values, and its $n$ neighbors, i.e., assignments at Hamming distance one from $s$. Of these neighbors, $j$ have $j - 1$ bad values, and the remaining $n - j$ have $j + 1$ bad values. For $j \leq n - k$, $s$ has $j$ neighbors with fewer conflicts and $n - j$ with more. Thus for these assignments, examining the number of conflicts in the neighbors readily determines the value of $j$. When $j = n - k + 1$, the assignment continues to have $j$ neighbors with fewer conflicts, but now the remaining $k - 1$ neighbors have the same number of conflicts since the additional bad value does not increase the number of conflicts. Finally, the neighbors of assignments with $n - k + 1 < j \leq n$ all have the same number of conflicts. Thus examining the number of conflicts in an assignment's neighbors determines the value of $j$, with the exception that assignments with $n - k + 1 < j \leq n$ are not distinguishable.

The value of $j$ is the number of conflicts that $s$ would have in the maximally constrained 1-SAT problem with the same solution as the given maximally constrained $k$-SAT problem. Thus for a maximally constrained $k$-SAT problem let

$$c_{\mathrm{eff}}(s) = \begin{cases} j & \text{if } j \leq n - k + 1 \\ n - k + 2 & \text{otherwise} \end{cases} \tag{7}$$

The value of $c_{\mathrm{eff}}(s)$ can be determined rapidly, in much the same way that a classical local search method checks the number of conflicts among neighbors of its current state to determine which assignment to move to next (Minton, Johnston, Philips, & Laird, 1992; Selman, Levesque, & Mitchell, 1992). Thus $c_{\mathrm{eff}}$ is a computationally tractable approximation to the number of conflicts each assignment would have in the corresponding 1-SAT problem. Except for a few assignments with many conflicts, $c_{\mathrm{eff}}$ gives the correct value. Specifically, only assignments with at least $n - k + 3$ bad values are given an incorrect value of $j$ by this approximation. In particular, the approximation is completely correct for $k = 2$.

While classical searches use the number of conflicts in an assignment and its neighbors, another possibility for maximally constrained problems is to use the number of conflicts in the assignment and its complement (i.e., the assignment with opposite values for all the variables). If the assignment has $j$ bad values, its complement will have $n - j$. As described above, the number of conflicts in an assignment uniquely determines the corresponding value of $j$ provided $j \leq n - k$. On the other hand, the number of conflicts in the complement assignment uniquely determines $j$ when $n - j \leq n - k$, or $j \geq k$. Thus, as long as $n > 2k$, at least one of these conditions will be true for all $0 \leq j \leq n$ and the correct value of $j$ can be determined.

## 3.2 Random SAT Problems

Theoretically, search algorithms are often evaluated for the worst possible case. However, in practice, search problems are often found to be considerably easier than suggested by these worst case analyses (Hogg, Huberman, & Williams, 1996). This observation leads to examining the typical behavior of search algorithms with respect to a specified *ensemble* of problems, i.e., a class of problems and a probability for each to occur. For instance, the ensemble of random $k$-SAT is specified by the number of variables $n$, the size of the clauses $k$ and the number of distinct[2] clauses $m$.

The quantum algorithm presented in this paper is effective for highly constrained soluble $k$-SAT problems. When there are many constraints, soluble problems are very rare among randomly generated instances. Thus to study the algorithm's behavior we generate random problems with a prespecified solution. That is, a random assignment is selected to be a solution and used to restrict the selection of clauses for the problem. In the remainder of this section we describe two methods for generating such problems, and how they can be related to corresponding 1-SAT problems.

### 3.2.1 Prespecified Solution

The most common use of a prespecified solution is to simply avoid selecting any clauses that conflict with it. Thus, we generate problems by randomly selecting a set of $m$ distinct clauses from among the $m_{\mathrm{max}}$, given by Eq. (5), available clauses (Nijenhuis & Wilf, 1978).

Consider a given soluble $k$-SAT problem with $m$ clauses, and let the assignment $r$ be one of its solutions. With respect to the solution $r$, we can define the bad value for each

---

2. This ensemble differs slightly from other studies where the clauses are not required to be distinct.

variable. For an assignment with $j$ bad values, the probability it has $c$ conflicts is

$$P_{\text{conf}}(c|j) = \frac{\binom{c_{\max}(j)}{c}\binom{m_{\max}-c_{\max}(j)}{m-c}}{\binom{m_{\max}}{m}} \tag{8}$$

where $c_{\max}(j)$, given by Eq. (6), is the largest possible number of conflicts for an assignment with $j$ bad values. The probability that an assignment has $j$ bad values is

$$P_{\text{bad}}(j) = 2^{-n}\binom{n}{j}$$

From these expressions and the definition of conditional probability, the probability that an assignment with $c$ conflicts has $j$ bad values is

$$P_{\text{bad}}(j|c) \propto P_{\text{conf}}(c|j)P_{\text{bad}}(j) \tag{9}$$

Hence, for a given assignment $s$ with $c$ conflicts, we can estimate the number of bad values it has by picking $j$ that maximizes $P_{\text{bad}}(j|c)$. We use this maximum likelihood value for $j$ as $c_{\text{eff}}(s)$ instead of Eq. (7) for random soluble $k$-SAT problems. This estimate is readily computed from the number of conflicts.

### 3.2.2 Balanced Clauses

Generating problems with a prespecified solution as described above is commonly used to study search problems. However, for each variable there are more allowable clauses where the variable is assigned its bad value than its good value. This makes highly constrained instances particularly easy since the good value for each variable can often be determined from its assigned value that appears most often in the clauses (Gent, 1998).

This bias in clause selection can be removed by a slight change in the generation method (Van Gelder & Tsuji, 1993). Specifically, instead of only avoiding those clauses that conflict with the prespecified solution, i.e., specify zero bad values, we also avoid any clauses that have an even number of bad values with respect to the prespecified solution. This selection method means both values for each variable appear equally often among the clauses. These balanced problems can have at most

$$m_{\max}^{\text{bal}} = \binom{n}{k}2^{k-1} \tag{10}$$

clauses. Furthermore, an assignment with $j$ bad values can have at most

$$c_{\max}^{\text{bal}}(j) = \sum_i{}' \binom{j}{i}\binom{n-j}{k-i} \tag{11}$$

conflicts, where the sum is over odd values of $i$. Using these values in Eq. (8) instead of $m_{\max}$ and $c_{\max}(j)$ gives the maximum likelihood estimate for $j$ in this "balanced clause" ensemble conditioned on the number of conflicts in the assignment.

## 4. Solving 1-SAT

A quantum computer, operating on superpositions of all assignments for *any* 1-SAT problem, can find a solution in a single search step (Hogg, 1998b). As a basis for solving highly constrained problems with larger clause sizes, we focus on maximally constrained 1-SAT which, from Eq. (5), has $m = n$ clauses and allows a simple specification. Specifically, we first motivate and define the algorithm for this case and illustrate it with small examples. Then we show that it is guaranteed to find a solution if one exists, and finally describe how the algorithm can be efficiently implemented on a quantum computer. The remainder of the paper then shows how this algorithm can form the basis for effectively solving highly constrained $k$-SAT for $k > 1$.

### 4.1 Motivation

Solving a search problem with a quantum computer requires finding a unitary operator $L$ that transforms an easily constructed initial state vector to a new state with large amplitude in those components of the state vector corresponding to solutions. Furthermore, determining this operator and evaluating it with the quantum computer must be tractable operations. This restriction means that any information used for a particular assignment must itself be easily computed, and the algorithm only uses readily computable unitary operations.

To design a single-step quantum algorithm, we consider superpositions of all assignments for the problem. Since we have no *a priori* knowledge of the solution, an unbiased initial state vector $\psi$ is one with equal amplitude in each assignment: $\psi_s = 2^{-n/2}$.

We must then incorporate information about the particular problem to be solved into this state vector. As in previous algorithms (Grover, 1997b; Hogg, 1996, 1998a), we do this by adjusting the phases in parallel, based on a readily computed property of each assignment: its number of conflicts with the constraints of the problem. This amounts to multiplication by a diagonal matrix $R$, with the entries on the diagonal having unit magnitude so that $R$ is unitary. The resulting amplitude for assignment $s$ is then of the form $\rho(s)2^{-n/2}$ where $|\rho(s)| = 1$ and $\rho(s)$ depends only on the number of conflicts in $s$. While this operation adds problem specific information to the state vector, in itself it does not solve the problem: at this point a measurement would return assignment $s$ with probability $|\rho(s)|^2 2^{-n} = 2^{-n}$, the same as random selection.

This operation also illustrates how the unitarity requirement, $|\rho(s)| = 1$, prevents us from using a computationally more desirable selection, i.e., $\rho(s) = 0$ if $s$ is not a solution, and nonzero otherwise. Such a choice, if possible, would immediately give a state vector with all amplitude in the solution. While determining whether a given assignment is a solution can be done rapidly for any NP problem, that information can not be directly used to set amplitudes of the nonsolutions to zero. Thus, while quantum parallelism allows rapid testing of all assignments, the restriction to unitary operators severely limits the *use* that can be made of this information.

For a single-step search algorithm, the remaining operations must not require any additional evaluation of the problem constraints, i.e., these operations will be the same for all problems of a given size. One the other hand, this restriction has the advantage of allowing more general unitary matrices than just phase adjustments. Specifically, this allows oper-ations that mix the various components of the state vector. We need to identify a mixing operator $U$ that makes all contributions to the solution add together in phase, but with $U$ independent of the particular problem.

The final result of the algorithm is $\phi_r = \sum_s U_{rs}\rho(s)2^{-n/2}$. Suppose $t$ is the solution. The maximum possible contribution to $\phi_t$ will be when all values in the sum combine in-phase. This will be the case if $U_{ts} = \rho^*(s)2^{-n/2}$ where $\rho^*$ is the complex conjugate of $\rho$. In this case, $\phi_t = \sum_s |\rho(s)|^2 2^{-n}$ which is just equal to 1. However, the mixing matrix itself is to be independent of any particular problem. Thus the issue is whether it is possible to create a $U$ whose values will have the required phases no matter where the solution is. One approach is to note that the mixing should have no bias as to the amount of amplitude that will need to be moved from one assignment to another in the state vector. This means that the magnitude of each element in $U$ will be the same, i.e., $|U_{rs}| = 2^{-n/2}$. For the phase of each element, we can consider using the feature of assignments used in classical local searches, namely the neighbors of each assignment. This suggests having $U_{rs}$ depend only on the Hamming distance between $r$ and $s$, i.e., $U_{rs} = 2^{-n/2}\mu_{d(r,s)}$ where $|\mu_d| = 1$.

With the elements of $U$ depending only on the Hamming distance, the matrix is independent of any particular problem's constraints. The question is then whether some feasible choices of $\rho(s)$ and $\mu_d$ allow $2^{-n}\sum_s \mu_{d(t,s)}\rho(s) = 1$ for the solution $t$. This will be the case provided $\mu_{d(t,s)} = \rho^*(s)$, where $\rho(s) = \rho_c$ depends only on the number of conflicts $c$ in assignment $s$. This relation does not hold for all search problems. However, for the maximally constrained 1-SAT considered here, the Hamming distance of assignment $s$ from the solution $t$, $d(t,s)$, which is the number of bad values in $s$, is precisely equal to the number of conflicts in $s$. Thus, to ensure all amplitude is combined into the solution, we merely need to have $\mu_d = \rho^*_d$.

The final question is what choices for the $\mu_d$ values are consistent with $U$ being a unitary matrix. This requirement restricts the available choices, e.g., having all $\mu_d = 1$ results in the nonunitary matrix with all elements equal to $2^{-n/2}$.

To examine the possible choices, consider the smallest possible case, $n = 1$. One maximally constrained, but still solvable, problem has the single clause NOT $V_1$ and solution $V_1 =$ false. The two assignments, 0 and 1, have, respectively, 0 and 1 conflicts. Since overall phase factors are irrelevant, we can select $\rho_0 = 1$ leaving a single remaining choice for $\rho_1$. For the matrix $U$, we have pairs of assignments with Hamming distance 0 and 1. Requiring $\mu_d = \rho^*_d$ then gives

$$U = \frac{1}{\sqrt{2}}\begin{pmatrix} 1 & \rho^*_1 \\ \rho^*_1 & 1 \end{pmatrix}$$

The unitarity condition, $U^\dagger U = I$, then requires that $\rho_1$ be purely imaginary, i.e., $\rho_1 = \pm i$. We arbitrarily pick $\rho_1 = i$. Starting from the initial state with equal amplitude in both assignments we then have the results of applying $R$ followed by $U$:

$$\frac{1}{\sqrt{2}}\begin{pmatrix} 1 \\ 1 \end{pmatrix} \to \frac{1}{\sqrt{2}}\begin{pmatrix} 1 \\ i \end{pmatrix} \to \begin{pmatrix} 1 \\ 0 \end{pmatrix}$$

giving all the amplitude in the solution. The overall operation $L = UR$ is

$$L = \frac{1}{2}\begin{pmatrix} 1 & 1 \\ -i & i \end{pmatrix}$$

It is important to note that the *same* operations also work if, instead, the other assignment is the solution, i.e., the problem has the clause $V_1$ and solution $V_1 = $ true. In this case, the assignments 0 and 1 now have, respectively, 1 and 0 conflicts so the $\rho_1 = i$ phase adjustment is now applied to assignment 0. The operation then gives

$$\frac{1}{\sqrt{2}} \begin{pmatrix} 1 \\ 1 \end{pmatrix} \rightarrow \frac{1}{\sqrt{2}} \begin{pmatrix} i \\ 1 \end{pmatrix} \rightarrow \begin{pmatrix} 0 \\ 1 \end{pmatrix}$$

Again, all amplitude is in the single solution, and the overall operation is

$$L = \frac{1}{2} \begin{pmatrix} i & -i \\ 1 & 1 \end{pmatrix}$$

While the overall operation $L$ depends on the location of the solution, for these problems it can be implemented by composing operators $U$ and $R$ that do not require knowledge of the solution. Instead, as described more generally in Section 4.5, $R$ is implemented by using the classical function for evaluating the number of conflicts in a given assignment, but applied to a superposition of all assignments.

With these motivating arguments for the form of the operations, examining a few larger values of $n$ establishes the simple pattern of phases used in the algorithm described in the remainder of this section.

## 4.2 The Algorithm for Maximally Constrained 1-SAT

Briefly, the algorithm starts with an equal superposition of all the assignments, adjusts the phases of the amplitudes based on the number of conflicts in the assignments, and then mixes the amplitudes from different assignments. This algorithm requires only a single testing of the assignments, corresponding to a single classical search step.

Specifically, the initial state is $\psi_s = 2^{-n/2}$ for each of the $2^n$ assignments $s$, and the final state vector is

$$\phi = UR\psi \tag{12}$$

where the matrices $R$ and $U$ are defined as follows. The matrix $R$ is diagonal with $R_{ss} \equiv \rho(s)$ depending on the number of conflicts $c$ in the assignment $s$, ranging from 0 to $n$:

$$\rho(s) = \rho_c = i^c \tag{13}$$

The mixing matrix elements $U_{rs} = u_{d(r,s)}$ depend only on the Hamming distance between the assignments $r$ and $s$, with

$$u_d = 2^{-n/2}(-i)^d \tag{14}$$

and $d$ ranging from 0 to $n$.

## 4.3 Examples

To illustrate the algorithm, consider the example problem in Section 3. It has $n = m = 2$. With the assignments ordered according to the corresponding interger value, i.e., 00, 01, 10, and 11, $U = A^{(2)}/2$ where

$$A^{(2)} = \begin{pmatrix} 1 & -i & -i & -1 \\ -i & 1 & -1 & -i \\ -i & -1 & 1 & -i \\ -1 & -i & -i & 1 \end{pmatrix} \tag{15}$$

The resulting behavior is:

| assignment $s$ | 00 | 01 | 10 | 11 |
|---|---|---|---|---|
| number of conflicts | 0 | 1 | 1 | 2 |
| $\rho(s)$ | 1 | $i$ | $i$ | $-1$ |
| $\psi$ | 1/2 | 1/2 | 1/2 | 1/2 |
| $R\psi$ | 1/2 | $i/2$ | $i/2$ | $-1/2$ |
| $\phi = UR\psi$ | 1 | 0 | 0 | 0 |

giving an amplitude of 1 in the solution assignment 00.

Another example, with $n = m = 3$ is the propositional formula (NOT $V_1$) AND (NOT $V_2$) AND $V_3$, with assignments $000, 001, 010, \ldots, 110, 111$, represented as bit vectors, and solution $\{V_1 = \text{false}, V_2 = \text{false}, V_3 = \text{true}\}$, i.e., the bit vector 100. In this case $U$ can be expressed in terms of $A^{(2)}$ from Eq. (15) in block form:

$$U = \frac{1}{\sqrt{8}} \begin{pmatrix} A^{(2)} & -iA^{(2)} \\ -iA^{(2)} & A^{(2)} \end{pmatrix} \qquad (16)$$

For this case, the algorithm's behavior is:

| assignment $s$ | 000 | 001 | 010 | 011 | 100 | 101 | 110 | 111 |
|---|---|---|---|---|---|---|---|---|
| number of conflicts | 1 | 2 | 2 | 3 | 0 | 1 | 1 | 2 |
| $\rho(s)$ | $i$ | $-1$ | $-1$ | $-i$ | 1 | $i$ | $i$ | $-1$ |
| $\sqrt{8}\psi$ | 1 | 1 | 1 | 1 | 1 | 1 | 1 | 1 |
| $\sqrt{8}R\psi$ | $i$ | $-1$ | $-1$ | $-i$ | 1 | $i$ | $i$ | $-1$ |
| $\phi = UR\psi$ | 0 | 0 | 0 | 0 | 1 | 0 | 0 | 0 |

Again, all the amplitude is in the solution.

## 4.4 Performance of the Algorithm

Consider a maximally constrained soluble 1-SAT problem with $n$ variables. As described in Section 3, in such a problem each clause involves a separate variable and there is exactly one solution. To show that the algorithm works for all $n$, we evaluate Eq. (12).

For each assignment $s$, $(R\psi)_s = \rho(s)2^{-n/2}$ from Eq. (13). Then for each assignment $r$, $(UR\psi)_r = 2^{-n/2} \sum_s U_{rs}\rho(s)$. Each $s$ in this sum can be characterized by

- $x$: the number of conflicts $s$ shares with $r$

- $y$: the number of conflicts of $s$ that are not conflicts of $r$

Let $h$ be the number of conflicts in the assignment $r$, i.e., the number of the $n$ variables to which $r$ assigns an incorrect value. In terms of these quantities, $s$ has $x + y$ conflicts and is at Hamming distance $d(r, s) = (h - x) + y$ from $r$. The number of such assignments is $\binom{h}{x}\binom{n-h}{y}$, so the sum can be written as

$$(UR\psi)_r = 2^{-n/2} \sum_x \binom{h}{x} \sum_y \binom{n-h}{y} u_{h-x+y}\rho_{x+y} \qquad (17)$$

Substituting the values from Eq. (13) and (14), gives

$$
\begin{aligned}
(UR\psi)_r &= 2^{-n} \sum_x \binom{h}{x} \sum_y \binom{n-h}{y} (-i)^{h-x+y} i^{x+y} \qquad (18) \\
&= 2^{-n} (-i)^h 2^{n-h} \sum_x \binom{h}{x} (-1)^x
\end{aligned}
$$

This gives $(UR\psi)_r = \delta_{h0}$ where $\delta_{xy} = 1$ if $x = y$ and 0 otherwise by use of the identity

$$
\sum_{x=0}^{h} (-1)^x \binom{h}{x} = \delta_{h0} \qquad (19)
$$

Thus, $\phi = UR\psi$ has all its amplitude in the state with no conflicts, i.e., the unique solution. A measurement made on this final state is guaranteed to produce a solution.

## 4.5 Implementation

Conceptually, the operation of Eq. (12) can be performed classically by matrix multiplication. However, since the matrices have $2^n$ rows and columns, this is not be a practical algorithm. As described in Section 2, quantum computers can rapidly perform many matrix operations of this size. Here we show how this is possible for the operations used by this algorithm.

For describing the implementation, it is useful to denote the individual components in a superposition explicitly. Traditionally, this is done using the ket notation introduced by Dirac (1958). For instance, the superposition described by the state vector of Eq. (1) is equivalently written as $\sum_s \psi_s |s\rangle$ where $|s\rangle$ just represents a unit basis vector corresponding to the assignment $s$. An example of these alternate, and equivalent, notations is:

$$
\begin{pmatrix} \psi_0 \\ \psi_1 \end{pmatrix} = \psi_0 \begin{pmatrix} 1 \\ 0 \end{pmatrix} + \psi_1 \begin{pmatrix} 0 \\ 1 \end{pmatrix} = \psi_0 |0\rangle + \psi_1 |1\rangle
$$

### 4.5.1 Forming the Initial Superposition

The initialization of $\psi$ can be performed rapidly by applying the matrix of Eq. (3) separately to each of the $n$ bits. For instance, when $n = 2$, starting from both bits set to 0, the state vector is changed as:

$$
\begin{aligned}
|00\rangle &\rightarrow \frac{1}{\sqrt{2}} \left( |00\rangle + |01\rangle \right) \\
&\rightarrow \frac{1}{2} \left( (|00\rangle + |10\rangle) + (|01\rangle + |11\rangle) \right)
\end{aligned}
$$

Equivalently, in terms of state vectors, this is

$$
\begin{pmatrix} 1 \\ 0 \\ 0 \\ 0 \end{pmatrix} \rightarrow \frac{1}{\sqrt{2}} \begin{pmatrix} 1 \\ 1 \\ 0 \\ 0 \end{pmatrix} \rightarrow \frac{1}{2} \begin{pmatrix} 1 \\ 1 \\ 1 \\ 1 \end{pmatrix}
$$

### 4.5.2 Adjusting Phases

For the operation $R\psi$, note that each value in Eq. (13) has unit magnitude so $R$ is a unitary diagonal matrix. Furthermore each $\rho_c$ only requires using the efficient classical procedure $f(s)$ that counts the number of conflicts in an assignment $s$. We require a reversible version of this procedure, which can be made with an additional program register. When the phases to be introduced are just $\pm1$, this additional register needs to take on only two values, 0 or 1, corresponding to whether the phase should be 1 or $-1$, respectively. Thus it can be represented with a single additional quantum bit, beyond those required to represent the assignment. Such phases have been used in previous algorithms (Hogg, 1996; Grover, 1997b) and can be implemented through a single evaluation of $f(s)$ by setting the extra variable to be a superposition of its two values (Boyer et al., 1996).

In the algorithm presented here, Eq. (13) requires phases that are powers of $i$, which can take on four different values: 1, $i$, $-1$ and $-i$. The technique used with $\pm1$ phases can be generalized to work with these four values, again with a single evaluation of $f(s)$. The additional register must consist of two quantum bits, so it can take on the values 0, 1, 2 or 3. For an assignment $s$ and register $x$, we use the reversible operation

$$F : |s, x\rangle \rightarrow |s, x + c \bmod 4\rangle$$

where $c = f(s)$ is the number of conflicts in assignment $s$. It then remains to show how this operation can be used to perform the required phase adjustments. Just as we operate with a superposition of all possible assignments, to implement the phase adjustment, we set register $x$ to be a particular superposition of its four values: $\Psi = \frac{1}{2}(|0\rangle - i|1\rangle - |2\rangle + i|3\rangle)$. One way to construct this superposition is to start with both bits of $x$ set to 1, operate on the most significant bit with Eq. (3) and then operate on the other bit with

$$\frac{1}{\sqrt{2}} \begin{pmatrix} -i & 1 \\ 1 & -i \end{pmatrix}$$

to get

$$
\begin{aligned}
|11\rangle &\rightarrow \frac{1}{\sqrt{2}}(|01\rangle - |11\rangle) \\
&\rightarrow \frac{1}{2}((|00\rangle - i|01\rangle) - (|10\rangle - i|11\rangle))
\end{aligned}
$$

This is just the superposition $\Psi$ when we make the correspondence between the 2-bit vectors $00, \ldots, 11$ and the integers $0, \ldots, 3$, respectively.

We start with the equal superposition of amplitudes for the assignments and this superposition for $x$:

$$2^{-n/2} \sum_s |s\rangle \Psi = 2^{-n/2-1} \sum_s \sum_{x=0}^{3} (-i)^x |s, x\rangle$$

As illustrated with Eq. (2), the operation $F$ acts on each term in this superposition separately, to produce

$$2^{-n/2-1} \sum_{sx} (-i)^x |s, x + c \bmod 4\rangle$$

where $c$ is the number of conflicts in assignment $s$. Let $y = x + c \bmod 4$. Then, for a given assignment $s$, as $x$ ranges from 0 to 3, $y$ also takes on these values, but not necessarily in the same order. Thus this resulting superposition can also be written as

$$2^{-n/2-1} \sum_s \sum_{y=0}^{3} (-i)^{y-c} |s, y\rangle$$

because $(-i)^4 = 1$. In this form, the sums separate to give finally

$$2^{-n/2} \sum_s i^c |s\rangle \sum_{y=0}^{3} (-i)^y |y\rangle = 2^{-n/2} \sum_s i^c |s\rangle \Psi$$

The net result of applying $F$ using the superposition $\Psi$ for the additional register is to change the phase of each assignment $s$ by $i^c$, as required by Eq. (13). Importantly, the final result reproduces the original factored form in which the superposition of assignments is not correlated with the superposition of the register. This factored form means the register plays no role in the subsequent mixing operation applied by the matrix $U$ to the superposition of assignments. Thus this procedure produces the required phase changes using only one evaluation of $f(s)$, showing how the phases of a superposition of assignments can be adjusted without requiring any prior explicit knowledge of the solution.

### 4.5.3 The Mixing Matrix

To implement $U$ specified by Eq. (14) we use two simpler matrices, $W$ and $\Gamma$ defined as follows. For assignments $r$ and $s$,

$$W_{rs} = 2^{-n/2}(-1)^{|r \wedge s|} \tag{20}$$

is the Walsh transform and $\Gamma$ is a diagonal matrix whose elements $\Gamma_{rr} \equiv \gamma(r)$ depend only on the number of 1-bits in each assignment, namely,

$$\gamma(r) = \gamma_h \equiv i^h e^{-i\pi n/4} \tag{21}$$

where $h = |r|$, ranging from 0 to $n$. The overall phase, $e^{-i\pi n/4}$, is not essential for the algorithm. It merely serves to make the final amplitude in the solution be one rather than $e^{i\pi n/4}$. Whether or not this overall phase is used, the probability to find a solution is one.

The matrix $W$ is unitary and can be implemented efficiently (Boyer et al., 1996; Grover, 1997b). For $n = 1$, $W$ is just the matrix of Eq. (3). The phases in the matrix $\Gamma$ are powers of $i$ and so can be computed rapidly using similar procedures to those described above for the matrix $R$. In this case we use a procedure that counts the number of 1-bits in each assignment rather than the number of conflicts.

Finally we show that $U$ can be implemented by the product $W\Gamma W$. To see this, let $\hat{U} \equiv W\Gamma W$. Then

$$\hat{U}_{rs} = 2^{-n} \sum_{h=0}^{n} \gamma_h S_h(r, s) \tag{22}$$

where

$$S_h(r, s) = \sum_{t, |t|=h} (-1)^{|r \wedge t| + |s \wedge t|} \tag{23}$$

with the sum over all assignments $t$ with $h$ 1-bits. Each 1-bit of $t$ contributes 0, 1 or 2 to $|r \wedge t| + |s \wedge t|$ when the corresponding positions of $r$ and $s$ are both 0, have exactly a single 1-bit, or are both 1, respectively. Thus $(-1)^{|r \wedge t| + |s \wedge t|}$ equals $(-1)^z$ where $z$ is the number of 1-bits in $t$ that are in exactly one of $r$ and $s$. There are $(|r| - |r \wedge s|) + (|s| - |r \wedge s|)$ positions from which such bits of $t$ can be selected, and by Eq. (4) this is just $d(r, s)$. Thus the number of assignments $t$ with $h$ 1-bits and $z$ of these bits in exactly one of $r$ and $s$ is given by $\binom{d}{z}\binom{n-d}{h-z}$ where $d = d(r, s)$. Thus $S_h(r, s) = S_{hd}^{(n)}$ where

$$S_{hd}^{(n)} = \sum_z (-1)^z \binom{d}{z}\binom{n-d}{h-z} \tag{24}$$

so that $\hat{U}_{rs} = \hat{u}_{d(r,s)}$ with $\hat{u}_d = 2^{-n} \sum_h \gamma_h S_{hd}^{(n)}$. Substituting the value of $\gamma_h$ from Eq. (21) then gives

$$
\begin{aligned}
\hat{u}_d &= 2^{-n} e^{-i\pi n/4} \sum_{hz} i^h (-1)^z \binom{d}{z}\binom{n-d}{h-z} \\
&= 2^{-n} e^{-i\pi n/4} (1-i)^d (1+i)^{n-d}
\end{aligned}
$$

which equals $u_d$ as defined in Eq. (14). Thus $U = WTW$, allowing $U$ to be efficiently implemented. As a final note, except for a different choice of the $\gamma_h$ phases, this is the same implementation as used for the mixing matrix defined in an unstructured quantum search algorithm (Grover, 1997b).

### 4.5.4 Required Search Time

While search algorithm performances are often compared based on the number of search *steps* required, i.e., the number of sequentially examined assignments, it is also important to compare the number of more elementary computational operations required. At the most fundamental level, these operations are logic operators on one or two bits at a time (e.g., the logical not or exclusive-or operations). As described above, the matrix operations and forming the initial state can be done with a series of $O(n)$ bit operations (Boyer et al., 1996).

The time required to count the number of conflicts in an assignment depends on data structures used to represent the problem. A single evaluation will be comparable for both the quantum and classical algorithms. For a SAT problem with $m$ clauses, examining each clause to see if it conflicts with a given assignment uses $O(m)$ tests. Each of these tests will, in turn, require comparing at least part of the clause to the assignment. Because the clauses in $k$-SAT are of fixed size, this gives an overall cost of $O(m)$ to evaluate the number of conflicts.

For local classical search, the number of conflicts in neighboring assignments will also be evaluated to determine which assignment should be examined at the next step of the search. Since neighbors differ by the value of only one variable, in fact it is only necessary to examine clauses that involve that variable to determine the difference in the number of conflicts between an assignment and one of its neighbors. This evaluation will thus require only $O(m/n)$ tests. Examining each, or a least a good portion, of the $n$ neighbors results in a total of $O(m)$ tests to find the next assignment. Selecting an initial assignment requires a value for each variable, a cost of $O(n)$.

Thus we can expect both algorithms to involve costs of $O(n + m)$ to evaluate a single search step. That is, the cost for a *single* search step is about the same for the quantum algorithm and classical searches when neighbors are examined. However, the quantum algorithm is able to examine the characteristics of all assignments in superposition while a classical search examines just one state, allowing the quantum algorithm to complete in just one step while the best classical methods for $k$-SAT require $O(n)$ steps. For the highly constrained $k$-SAT problems with $k > 1$, discussed below, $m \gg n$ so the dominant cost will be in the evaluation of the number of conflicts.

This discussion indicates how a comparison of search steps gives a reasonable comparison in terms of elementary operations as well. However, a full comparison will also depend on details of actual implementations, such as any additional operations required for controlling errors that cannot themselves be performed in parallel with the higher level steps of the algorithm. These remain significant issues in the development of quantum computation (Landauer, 1994; Unruh, 1995; Haroche & Raimond, 1996; Monroe & Wineland, 1996), but at this point seem unlikely to be fundamental difficulties (Berthiaume, Deutsch, & Jozsa, 1994; Shor, 1995; Knill, Laflamme, & Zurek, 1998). In particular, because the algorithm requires only a single step, decoherence is likely to be less of a difficulty for it than search algorithms that require multiple repeated steps to move significant amplitude to solutions (Grover, 1997b; Hogg, 1998a).

Finally, search algorithms can be compared based on elapsed execution time. Current hardware implementations (Barenco, Deutsch, & Ekert, 1995; Bouwmeester et al., 1997; Chuang, Vandersypen, Zhou, Leung, & Lloyd, 1998; Cirac & Zoller, 1995; Cory, Fahmy, & Havel, 1996; Gershenfeld, Chuang, & Lloyd, 1996; Sleator & Weinfurter, 1995) are quite limited in size, so such a comparison will need to await the construction of quantum machines with a sufficient number of bits to perform interesting searches.

## 5. Applying the Algorithm to $k$-SAT

The algorithm presented above is effective for 1-SAT by exploiting the simple structure of such problems. As described in Section 3, many highly constrained $k$-SAT problems have a similar structure. This observation allows the 1-SAT algorithm to be applied to more general problems, although with a reduction in performance. Specifically, applying Eq. (13) requires knowing the number of conflicts the assignments would have in the corresponding maximally constrained 1-SAT problem whose solution is equal to one of the solutions of the original $k$-SAT problem. As described in Section 3.1, for the most part this can be computed efficiently using the neighborhood relations for the problem. This suggests simply changing the 1-SAT algorithm to use $\rho(s) = \rho_{c_{\text{eff}}(s)}$.

To see how this approximation changes the performance, consider an assignment $s$ with $y$ bad values with respect to a specific solution $r$ and let

$$\delta(s) = \rho_{c_{\text{eff}}(s)} - \rho_y \tag{25}$$

The vector $v^{(k)} = R\psi$ used with the $k$-SAT problem is related to the vector from the corresponding 1-SAT problem $v^{(1)}$ by $v_s^{(k)} = v_s^{(1)} + \delta_s$. Except for this change, the remaining transformations of the algorithm are the same as in the 1-SAT case. Thus

$$UR\psi = \phi^{(1)} + U\Delta\psi$$

where $\phi^{(1)}$ is the result of the corresponding 1-SAT problem, i.e., all amplitude in the solution, and $\Delta$ is a diagonal matrix, with elements given by $\delta(s)$. It is convenient to define the average value of $\delta(s)$ over all assignments with $y$ bad values:

$$\delta_y = \frac{1}{\binom{n}{y}}\sum_s{}' \delta(s) \qquad (26)$$

where the sum is restricted to just the $\binom{n}{y}$ assignments with $y$ bad values.

The change in the amplitude in the solution state is determined by $\eta \equiv (U\Delta\psi)_r$ when $r$ is the solution. This change can be expressed using Eq. (17) by recalling that $h = 0$ for the solution and replacing the phases $\rho_y$ by the error in the phases, $\delta_y$:

$$\eta = 2^{-n/2}\sum_{y=0}^{n}\binom{n}{y}u_y\delta_y \qquad (27)$$

Since all the $\rho_y$ have unit magnitude, $|\delta_y| \leq 2$. If the problem has only one solution, the probability the algorithm will find it is $P_{\text{soln}} = |1 + \eta|^2$. If there are multiple solutions, this is a lower bound on the probability.

The following sections use these observations to extend the range of problems to which the 1-SAT algorithm can be effectively applied.

## 6. Solving Maximally Constrained $k$-SAT

The regular structure of maximally constrained soluble $k$-SAT problems allows them to be solved in $O(1)$ steps. That is, the probability to find a solution remains $O(1)$ as $n$ increases. Thus a solution is very likely to be found by repeating the algorithm $O(1)$ times, and, as described above, each trial of the algorithm involves only one evaluation of the conflicts. To see this, we use $c_{\text{eff}}(s)$ from Eq. (7). For $k > 2$, this approximation results in incorrect phase choices for only a few, high-conflict assignments. Because the proportion of incorrect phases is small, we can expect this approximation will introduce only small amplitudes in nonsolution states. However, it will also make the algorithm incomplete: it can find a solution if one exists but not prove no solutions exist.

Specifically, $\delta(s)$ can be nonzero only for $y \geq n - k + 3$, where $y$ is the number of bad values in assignment $s$. Thus using Eq. (27), $|\delta_y| \leq 2$ and Eq. (14),

$$|\eta| \leq 2^{-n}2\sum_{y=n-k+3}^{n}\binom{n}{y}$$

When $n - k + 3 \geq n/2$, the sum over binomial coefficients can be bounded (Palmer, 1985) to give

$$|\eta| \leq 2^{-(n-1)}\binom{n}{k-3}\frac{n+1-(k-3)}{n+1-2(k-3)} \sim 2^{-(n-1)}\frac{n^{k-3}}{(k-3)!} \qquad (28)$$

Thus the probability to obtain a solution is

$$P_{\text{soln}} = |1 + \eta|^2 \geq (1 - |\eta|)^2 \sim 1 - 2^{-(n-2)} \frac{n^{k-3}}{(k-3)!} \rightarrow 1 \qquad (29)$$

which rapidly approaches 1. Hence, this algorithm is able to find the solution in $O(1)$ search steps as $n$ increases. This behavior is illustrated in Figure 1.

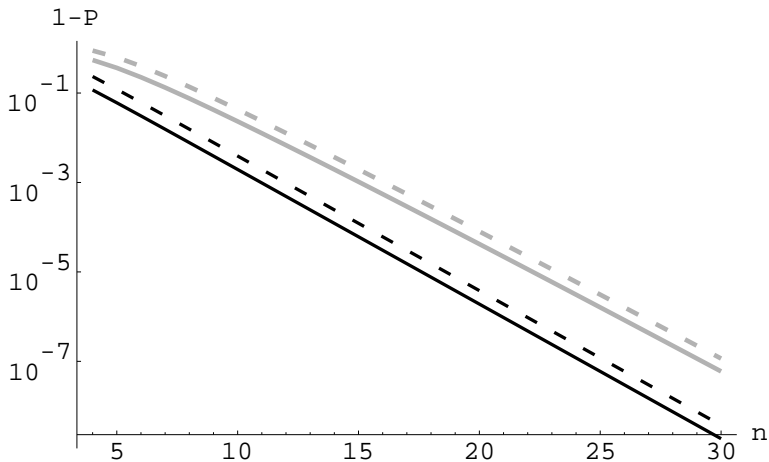

Figure 1: Behavior of $1 - P_{\text{soln}}$ vs. $n$ for maximally constrained soluble $k$-SAT for $k = 3$ (black) and 4 (gray). For comparison, the bounds $(1 - |\eta|)^2$ from Eq. (28) are shown as the dashed lines.

Similarly, soluble balanced $k$-SAT problems with the maximum possible number of clauses, given by Eq. (10), give good performance as shown in Figure 2. The behavior in this case is rather irregular and continues for larger values of $n$, but still gives a high probability to find a solution. For odd $k$, the probability for a solution is exactly one for many values of $n$. In fact, by including neighborhood information, the errors in the remaining cases can also be eliminated, giving a perfect algorithm for these problems. For even $k$, the balanced clauses force the problem to have two solutions with opposite values. Even though this problem structure differs significantly from that of a 1-SAT problem with a single solution, the algorithm is able to find solutions for $k = 4$ with probability of about $1/2$, even as $n$ increases.

## 7. Solving Highly Constrained Random $k$-SAT

The discussion of Section 3.2 shows how a maximum likelihood estimate for $c_{\text{eff}}$ can be computed for each assignment. This value can then be used to extend the algorithm to arbitrary $k$-SAT problems. To the extent that the errors introduced by this approximation are small, the quantum algorithm will have a substantial probability to find a solution in a single step.

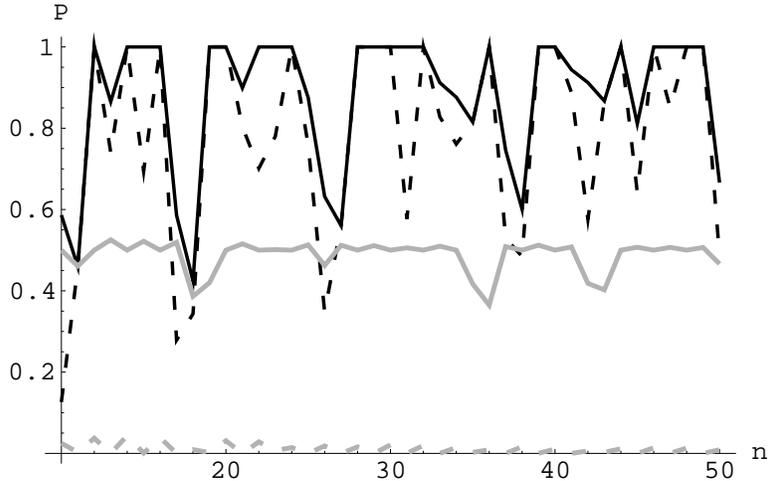

Figure 2: Behavior of $P_{\text{soln}}$ vs. $n$ for maximally constrained balanced soluble $k$-SAT for $k = 3$ (black) and $4$ (gray). For comparison, the bounds $(1 - |\eta|)^2$ based on Eq. (27) are shown as the dashed lines, and is quite small for the $k = 4$ case.

When averaged over the problem ensemble, the error given by Eq. (27) becomes

$$\overline{|\eta|} \leq B \equiv 2^{-n} 2 \sum_{y=0}^{n} \binom{n}{y} p_y$$

where $p_y$ is the probability an assignment with $y$ bad values is (incorrectly) determined to have a different number of bad values. In terms of the conditional probabilities of Section 3.2,

$$p_y = 1 - \sum_c P_{\text{conf}}(c|y) \chi_{cy}$$

where $\chi_{cy} = 1$ when the maximum likelihood estimate for a state with $c$ conflicts is $y$ (i.e., $c_{\text{eff}} = y$), and 0 otherwise.

For simplicity, these maximum likelihood estimates are determined solely from the number of conflicts in each state. The $c_{\text{eff}}$ values could be made a bit more accurate by including neighborhood information, as was used for maximally constrained random problems in Section 6.

Because highly constrained random SAT problems are relatively easy, they have not been well-studied with classical algorithms. Hence, to provide comparison with the quantum search results presented below, these problems were also solved with the classical GSAT procedure (Selman et al., 1992), limiting each trial to use no more than $2n$ steps before a new random initial state was selected. For both random and balanced ensembles, the median number of search steps required to find a solution grows linearly over the range of sizes considered here when $m = O(n^2)$. In particular, while the balanced ensemble has larger search costs, it still grows linearly when there is such a large number of clauses.

## 7.1 Random $k$-SAT

For random $k$-SAT with prespecified solution, Stirling's asymptotic expansion in Eq. (8) shows that $p_y = O(1)$ for $y = O(n)$ when $m$ grows as $O(n^2)$, which is much less that the maximum $m_{max} = O(n^k)$. In this case, the asymptotic behavior of the bound $B$ is determined by the values near the maximum of the binomial coefficient, i.e., near $y = n/2$. Thus if $m = \mu n^2$, we have $B \sim 2p_{n/2}$. From Eq. (29), $P_{soln} = O(1)$ at least when $B < 1$. This is the case for $\mu > \mu_{crit}$ where

$$\mu_{crit} = \begin{cases} 27\zeta^2/(2 + 18\zeta^2) & \text{if } k = 2 \\ 2(2^k - 1)^3\zeta^2/k^2 & \text{if } k > 2 \end{cases} \qquad (30)$$

where $\zeta \equiv \text{erf}^{-1}(\frac{1}{2}) \approx 0.477$. For instance, $\mu_{crit}$ is 1.01 and 17.3 for $k = 2$ and 3, respectively. Thus, the algorithm presented here is simple enough to allow an analytic bound on its behavior for highly constrained problems, thus demonstrating its asymptotic effectiveness in these cases. Other, more complex, structured quantum algorithms have only been evaluated empirically (Hogg, 1996, 1998a), which is limited to small problems.

The algorithm's behavior with fewer constraints, i.e., $\mu < \mu_{crit}$, is not easily evaluted analytically since the bound provided by $B$ is no longer useful. Instead, the behavior can be examined empirically using a classical simulation (Hogg & Yanik, 1998), which is however limited to problems with a relatively small number of variables. These empirical studies may eventually be extendable to larger problems using approximate evaluation techniques (Cerf & Koonin, 1997).

An example is given in Figure 3. This shows good performance for highly constrained problems, as expected from the behavior of the lower bound. Performance is also good with few constraints; not because the algorithm is capturing the problem structure particularly well but rather because there are many solutions to weakly constrained problems. As with other classical and quantum search methods that use problem structure, the hardest cases are for problems with an intermediate number of constraints (Hogg et al., 1996).

From Figure 4 we see that the nonzero asymptotic limit for the probability of a solution appears to continue for somewhat fewer constraints than expected from the value of $\mu_{crit}$. Below this value, the probability for finding a solution appears to decrease as a power of $n$, indicated by linear scaling on the log-log plot of Figure 4 for $m = n^2$. Similar empirical evaluations of the scaling with an intermediate number of constraints where the hard problems are concentrated, e.g., $m = 4n$ for 3-SAT, shows linear scaling on a log-plot, indicating exponential decrease in the probability to find a solution. Moreover, the resulting search costs in these cases are larger than those of other structured quantum search algorithms (Hogg, 1996, 1998a). Thus the structure of these harder cases differs enough from the simple 1-SAT problems that this algorithm is not effective for them.

In summary, the algorithm solves highly constrained problems with $m = \mu n^2$ in $O(1)$ steps for $\mu > \mu_{crit}$, and possibly for somewhat smaller values of $\mu$ as well. As the number of clauses is further reduced, the required number of steps appears to grow polynomially when $\mu > 0$ and exponentially when $m = O(n)$.

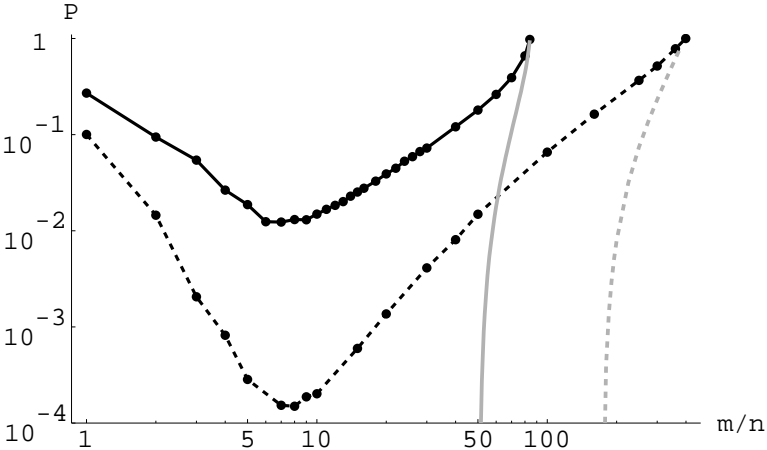

Figure 3: Probability to find a solution for random 3-SAT for $n = 10$ (solid) and 20 (dashed) vs. $m/n$, on a log-log scale. Each point is an average over at least 100 problem instances, and includes error bars for the standard deviation in this estimate of the averge. The error bars are smaller than the size of the plotted points. The gray lines show the corresponding lower bounds $(1 - B)^2$.

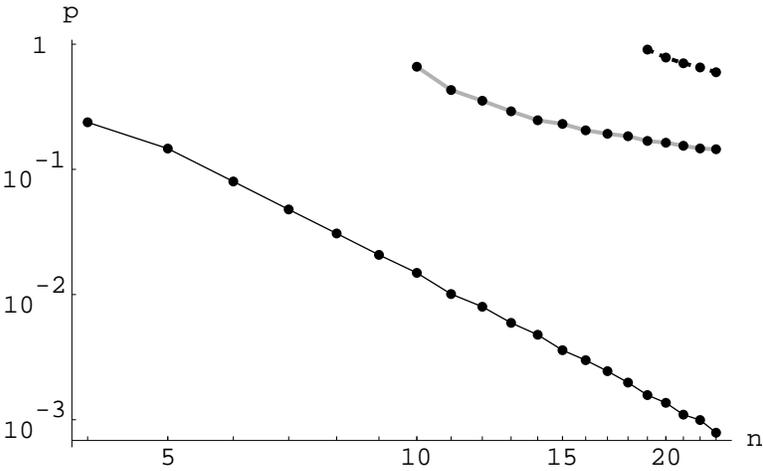

Figure 4: Probability to find a solution for random soluble 3-SAT vs. $n$ with, from top to bottom, $m = 18n^2$, $8n^2$ and $n^2$, respectively, on a log-log plot. For each value of $n$, at least 100 problem instances were used. Error bars showing the expected error in the estimate are included but are smaller than the size of the plotted points.

## 7.2 Balanced Clauses

In a similar way, the behavior of problems with balanced clauses can be evaluated, as shown in Figure 5. In this case the lower bound is much looser than for random soluble problems.

This is because, unlike the previous case, significant errors are made in assigning $c_{\text{eff}}$ for the large number of assignments with about $n/2$ bad values. The bound assumes that any such mistake gives the maximum possible contribution to Eq. (27), but in fact because of the limited phase choices in Eq. (13), some such mistakes will nevertheless give the correct value of the phase. Again, the intermediate problems are the most difficult for this algorithm.

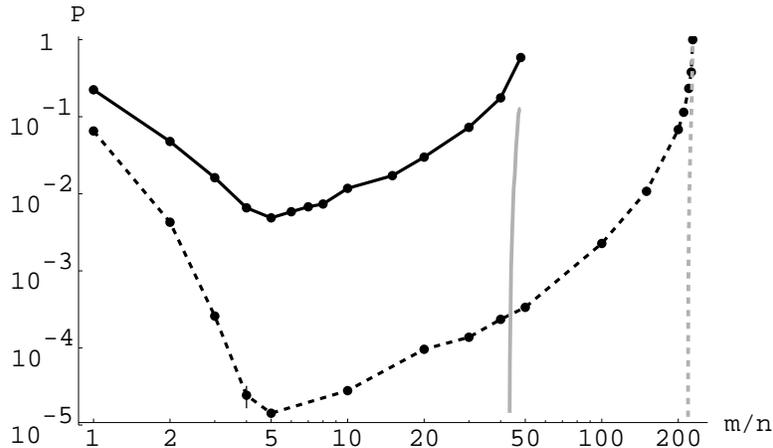

Figure 5: Probability to find a solution for random balanced 3-SAT for $n = 10$ (solid) and 20 (dashed) vs. $m/n$, on a log-log scale. Each point represents 100 problem instances and includes error bars which, in most cases, are smaller than the size of the plotted points. The gray lines show the corresponding lower bounds $(1 - B)^2$

Because the bound is so poor, its asymptotic behavior does not offer a useful guide to the behavior of the algorithm for highly constrained problems. Instead, the scaling for $m = O(n^2)$ is illustrated in Figure 6. The behavior is consistent with a polynomial decrease in the probability to find a solution, but definitive statements cannot be made from such small problem sizes.

## 8. Discussion

The algorithm presented in this paper provides an analytic demonstration that quantum machines can significantly exploit the structure of highly constrained $k$-SAT problems, thereby extending the range of search problems that definitely have effective quantum algorithms. This contrasts with previous work on structured quantum algorithms that could only be evaluated empirically.

In addition, for maximally constrained 2-SAT problems and many maximally constrained balanced $k$-SAT problems, the algorithm finds a solution with probability one. Thus in these cases, failure to find a solution definitely indicates the problem is not soluble, i.e., the search method is *complete*. As described in Section 3.1, with the slight additional cost of evaluating the number of conflicts in the complement of an assignment as well as the assignment itself for maximally constrained $k$-SAT, the correct corresponding 1-SAT problem can be determined. This additional information thus gives a complete search algo-rithm for maximally constrained $k$-SAT. This contrasts with previously proposed quantum algorithms that find solutions with probability less than one and hence cannot guarantee no solutions exist.

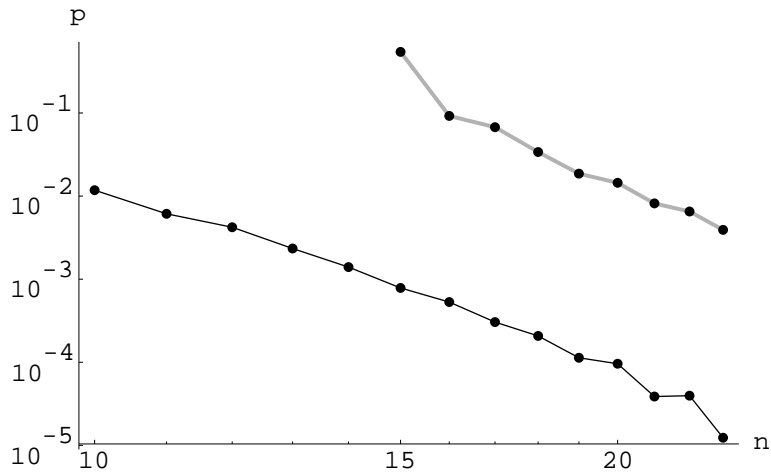

Figure 6: Probability to find a solution for random balanced 3-SAT vs. $n$ with, from top to bottom, $m = 8n^2$ and $n^2$, respectively, on a log-log plot. For each value of $n$, 100 problem instances were used. Error bars showing the expected error in the estimate are included but are smaller than the size of the plotted points.

One direction for future work is generalizing this algorithm to other types of combinatorial search. For instance, the algorithm is restricted to CSPs with two values per variable, such as SAT. While other CSPs can be recast as satisfiability problems, this mapping may obscure structure inherent in the original formulation. Thus it would be useful to find algorithms that apply directly to more general CSP formulations. One possible approach would be based on search methods that construct solutions incrementally from smaller parts, i.e., expanding the set of states to include assignments that give values to only some of the variables in the problem. Such a representation can apply readily to CSPs with any number of values for the variables (Hogg, 1996, 1998a). Another approach would examine replacing Walsh transforms with the approximate Fourier transform (Kitaev, 1995) as has been proposed to extend an unstructured search method to cases where the size of the search space is not a power of two (Boyer et al., 1996). Beyond CSPs, it would be interesting to investigate optimization problems.

Search problems with an intermediate number of constraints are the most difficult for classical heuristics as well as structure-based quantum searches (Hogg, 1996, 1998a) based on analogies with these classical methods. For $k$-SAT, these hard cases occur when the number of clauses grows linearly with the number of variables, which is much smaller than the $O(n^2)$ used in Section 7. The intermediate number of clauses creates considerable variance in the detailed structure of the search space from one problem instance to another. Thus one cannot rely on precise *a priori* knowledge of the structure in designing the algorithm. Nevertheless, the average or typical structure of these harder search problems has been characterized (Cheeseman, Kanefsky, & Taylor, 1991; Hogg et al., 1996; Hogg & Williams, 1994; Kirkpatrick & Selman, 1994; Monasson & Zecchina, 1996; Williams & Hogg, 1994) and may be suitable as a basis for developing appropriate search methods. Instead of aiming for a single-step algorithm, the large variation in structure is likely to require a series of smaller changes to the amplitudes, along with repeated tests of the consistency of all assignments, as with previous proposals (Grover, 1997b; Hogg, 1998a). However, since a quantum algorithm can explore all search paths simultaneously, it can avoid some of the variability encountered in classical methods: namely, that due to random selection of initial states or random tie-breaking when evaluating heuristics. Thus a quantum algorithm can focus on variation due only to differences in problem instances rather than also to the particular choices made in exploring a single search path. Ultimately, this observation may allow quantum algorithms to more usefully exploit improved understanding of typical problem structure than is feasible for classical methods.

This discussion raises the general issue of optimally using the information that can be readily extracted from CSP search states, as commonly used in classical heuristics. Such information includes the number of conflicts a state has and how it compares with its neighbors. Additional information is available on partial assignments, as used with incremental searches, but at the cost of involving a greatly expanded search space. The approach described in Section 5 suggests a useful technique is matching quantum algorithms to the average structure of search problem ensembles.

The most significant open question is the extent to which quantum algorithms can solve problems in polynomial time that require exponential time classically. Factoring provides one example (Shor, 1994) if, as commonly believed, it cannot be done in polynomial time classically. By contrast, highly constrained searches can be solved in polynomial time by both classical heuristics and, as shown in this paper, quantum machines. At the other extreme, searches that ignore problem structure are exponential, requiring $O(2^n)$ steps classically, and $O(2^{n/2})$ steps on quantum computers (Boyer et al., 1996). These observations can be summarized as:

| | cost scaling | |
| --- | --- | --- |
| type of problem | classical | quantum |
| unstructured | exponential | exponential |
| factoring | exponential | **polynomial** |
| highly constrained | **polynomial** | **polynomial** |

This comparison suggests some problems, including factoring, have enough structure to allow quantum machines to operate in polynomial time but not enough for classical machines to do so. Identifying the class of such problems is an important research direction for quantum computation. For example, an interesting open question is whether there is a scaling regime for the number of clauses, $m$, as a function of $n$ where the probability of finding a solution with a quantum machine decreases only polynomially with $n$ while classical searches require an exponential number of steps, even with the best known heuristics. This question is difficult to treat empirically, pointing to the need for further analytic investigation of quantum algorithms and the structure of search problems.

## Acknowledgments

I have benefited from discussions with Ian Gent, Carlos Mochon, Wolf Polak and Eleanor Rieffel.

## References

Barenco, A., Deutsch, D., & Ekert, A. (1995). Conditional quantum dynamics and logic gates. *Physical Review Letters, 74*, 4083–4086.

Benioff, P. (1982). Quantum mechanical hamiltonian models of Turing machines. *J. Stat. Phys., 29*, 515–546.

Bennett, C. H., & Landauer, R. (1985). The fundamental physical limits of computation. *Scientific American*, July, 48–56.

Bernstein, E., & Vazirani, U. (1993). Quantum complexity theory. In *Proc. 25th ACM Symp. on Theory of Computation*, pp. 11–20.

Berthiaume, A., Deutsch, D., & Jozsa, R. (1994). The stabilization of quantum computations. In *Proc. of the Workshop on Physics and Computation (PhysComp94)*, pp. 60–62 Los Alamitos, CA. IEEE Press.

Bouwmeester, D., et al. (1997). Experimental quantum teleportation. *Nature, 390*, 575–579.

Boyer, M., Brassard, G., Hoyer, P., & Tapp, A. (1996). Tight bounds on quantum searching. In Toffoli, T., et al. (Eds.), *Proc. of the Workshop on Physics and Computation (PhysComp96)*, pp. 36–43 Cambridge, MA. New England Complex Systems Institute.

Brassard, G., Hoyer, P., & Tapp, A. (1998). Quantum counting. Los Alamos preprint quant-ph/9805082.

Cerf, N. J., & Koonin, S. E. (1997). Monte Carlo simulation of quantum computation. In *Proc. of IMACS Conf. on Monte Carlo Methods*. Los Alamos preprint quant-ph/9703050.

Cerf, N. J., Grover, L. K., & Williams, C. P. (1998). Nested quantum search and NP-complete problems. Los Alamos preprint quant-ph/9806078.

Cerny, V. (1993). Quantum computers and intractable (NP-complete) computing problems. *Physical Review A, 48*, 116–119.

Cheeseman, P., Kanefsky, B., & Taylor, W. M. (1991). Where the really hard problems are. In Mylopoulos, J., & Reiter, R. (Eds.), *Proceedings of IJCAI91*, pp. 331–337 San Mateo, CA. Morgan Kaufmann.

Chuang, I. L., Vandersypen, L. M. K., Zhou, X., Leung, D. W., & Lloyd, S. (1998). Experimental realization of a quantum algorithm. *Nature, 393*, 143–146. Los Alamos preprint quant-ph/9801037.

Cirac, J. I., & Zoller, P. (1995). Quantum computations with cold trapped ions. *Physical Review Letters, 74*, 4091–4094.

Cory, D. G., Fahmy, A. F., & Havel, T. F. (1996). Nuclear magnetic resonance spectroscopy: An experimentally accessible paradigm for quantum computing. In Toffoli, T., et al. (Eds.), *Proc. of the Workshop on Physics and Computation (PhysComp96)*, pp. 87–91 Cambridge, MA. New England Complex Systems Institute.

Deutsch, D. (1985). Quantum theory, the Church-Turing principle and the universal quantum computer. *Proc. R. Soc. London A, 400*, 97–117.

Deutsch, D. (1989). Quantum computational networks. *Proc. R. Soc. Lond., A425*, 73–90.

Dirac, P. A. M. (1958). *The Principles of Quantum Mechanics* (4th edition). Oxford.

DiVincenzo, D. P. (1995). Quantum computation. *Science, 270*, 255–261.

Feynman, R. P. (1986). Quantum mechanical computers. *Foundations of Physics, 16*, 507–531.

Feynman, R. P. (1985). *QED: The Strange Theory of Light and Matter*. Princeton Univ. Press, NJ.

Feynman, R. P. (1996). *Feynman Lectures on Computation*. Addison-Wesley, Reading, MA.

Garey, M. R., & Johnson, D. S. (1979). *Computers and Intractability: A Guide to the Theory of NP-Completeness*. W. H. Freeman, San Francisco.

Gent, I. (1998). On the stupid algorithm for satisfiability. Tech. rep. APES-03-1998, Strathclyde University. Available at www.cs.strath.ac.uk/~apes/apereports.html.

Gershenfeld, N., Chuang, I., & Lloyd, S. (1996). Bulk quantum computation. In Toffoli, T., et al. (Eds.), *Proc. of the Workshop on Physics and Computation (PhysComp96)*, p. 134 Cambridge, MA. New England Complex Systems Institute.

Grover, L. K. (1997a). Quantum computers can search arbitrarily large databases by a single query. Los Alamos preprint quant-ph/9706005, Bell Labs.

Grover, L. K. (1997b). Quantum mechanics helps in searching for a needle in a haystack. *Physical Review Letters, 78*, 325–328.

Haroche, S., & Raimond, J.-M. (1996). Quantum computing: Dream or nightmare?. *Physics Today, 49*, 51–52.

Hogg, T. (1996). Quantum computing and phase transitions in combinatorial search. *J. of Artificial Intelligence Research, 4*, 91–128. Available at http://www.jair.org/abstracts/hogg96a.html.

Hogg, T. (1998a). A framework for structured quantum search. *Physica D, 120*, 102–116. Los Alamos preprint quant-ph/9701013.

Hogg, T. (1998b). Highly structured searches with quantum computers. *Physical Review Letters*, *80*, 2473–2476. Preprint at publish.aps.org/eprint/gateway/eplist/aps1997oct30_002.

Hogg, T., Huberman, B. A., & Williams, C. (1996). Phase transitions and the search problem. *Artificial Intelligence*, *81*, 1–15.

Hogg, T., & Williams, C. P. (1994). The hardest constraint problems: A double phase transition. *Artificial Intelligence*, *69*, 359–377.

Hogg, T., & Yanik, M. (1998). Local search methods for quantum computers. Los Alamos preprint quant-ph/9802043, Xerox.

Hoyer, P. (1997). Efficient quantum transforms. Los Alamos preprint quant-ph/9702028.

Kirkpatrick, S., & Selman, B. (1994). Critical behavior in the satisfiability of random boolean expressions. *Science*, *264*, 1297–1301.

Kitaev, A. Y. (1995). Quantum measurements and the Abelian stabilizer problem. Los Alamos preprint quant-ph/9511026.

Knill, E., Laflamme, R., & Zurek, W. H. (1998). Resilient quantum computation. *Science*, *279*, 342–345.

Landauer, R. (1994). Is quantum mechanically coherent computation useful?. In Feng, D. H., & Hu, B.-L. (Eds.), *Proc. of the Drexel-4 Symposium on Quantum Nonintegrability* Boston. International Press.

Lloyd, S. (1993). A potentially realizable quantum computer. *Science*, *261*, 1569–1571.

Mackworth, A. (1992). Constraint satisfaction. In Shapiro, S. (Ed.), *Encyclopedia of Artificial Intelligence*, pp. 285–293. Wiley, NY.

Minton, S., Johnston, M. D., Philips, A. B., & Laird, P. (1992). Minimizing conflicts: A heuristic repair method for constraint satisfaction and scheduling problems. *Artificial Intelligence*, *58*, 161–205.

Monasson, R., & Zecchina, R. (1996). The entropy of the k-satisfiability problem. *Phys. Rev. Lett.*, *76*, 3881–3885.

Monroe, C., & Wineland, D. (1996). Future of quantum computing proves to be debatable. *Physics Today*, *49*, 107–108.

Nijenhuis, A., & Wilf, H. S. (1978). *Combinatorial Algorithms for Computers and Calculators* (2nd edition). Academic Press, New York.

Palmer, E. M. (1985). *Graphical Evolution: An Introduction to the Theory of Random Graphs*. Wiley Interscience, NY.

Selman, B., Levesque, H., & Mitchell, D. (1992). A new method for solving hard satisfiability problems. In *Proc. of the 10th Natl. Conf. on Artificial Intelligence (AAAI92)*, pp. 440–446 Menlo Park, CA. AAAI Press.

Shor, P. (1995). Scheme for reducing decoherence in quantum computer memory. *Physical Review A*, *52*, 2493–2496.

Shor, P. W. (1994). Algorithms for quantum computation: Discrete logarithms and factoring. In Goldwasser, S. (Ed.), *Proc. of the 35th Symposium on Foundations of Computer Science*, pp. 124–134 Los Alamitos, CA. IEEE Press.

Sleator, T., & Weinfurter, H. (1995). Realizable universal quantum logic gates. *Physical Review Letters*, *74*, 4087–4090.

Terhal, B. M., & Smolin, J. A. (1997). Single quantum querying of a database. Los Alamos preprint quant-ph/9705041 v4, IBM.

Unruh, W. G. (1995). Maintaining coherence in quantum computers. *Physical Review A*, *51*, 992–997.

Van Gelder, A., & Tsuji, Y. K. (1993). Incomplete thoughts about incomplete satisfiability problems. In *Proc. of 2nd DIMACS Challenge*.

Williams, C. P., & Hogg, T. (1994). Exploiting the deep structure of constraint problems. *Artificial Intelligence*, *70*, 73–117.